\documentclass[conference]{IEEEtran}

\def\IEEEtitletopspace{18pt}

\usepackage{etoolbox}
\makeatletter
\patchcmd{\@makecaption}
  {\scshape}
  {}
  {}
  {}
\makeatother

\IEEEoverridecommandlockouts
\usepackage{cite}
\usepackage{amsmath,amssymb,amsfonts,gensymb}
\usepackage{algorithmic}
\usepackage{graphicx}
\usepackage{textcomp}
\usepackage{pifont}
\def\BibTeX{{\rm B\kern-.05em{\sc i\kern-.025em b}\kern-.08em
    T\kern-.1667em\lower.7ex\hbox{E}\kern-.125emX}}

\usepackage[table]{xcolor}
\usepackage{makecell}
\usepackage{siunitx}
\usepackage{multirow}

\graphicspath{{./sections/figs/}}

\title{VisTaNet: Attention Guided Deep Fusion for Surface Roughness Classification\\

}

\author{Prasanna Kumar Routray, Aditya Sanjiv Kanade, Jay Bhanushali, Manivannan Muniyandi

\thanks{Touch Lab, Center for Virtual Reality and Haptics, Indian Institute of Technology Madras, India}

\thanks{(Prasanna Kumar Routray, Aditya Sanjiv Kanade and Jay Bhanushali contributed equally to this work.) (Corresponding author: Prasanna Kumar Routray; prasanna.routray97@gmail.com)}
}

\begin{document}

\maketitle

\begin{abstract}
Human texture perception is a weighted average of multi-sensory inputs: visual and tactile. While the visual sensing mechanism extracts global features, the tactile mechanism complements it by extracting local features. The lack of coupled visuotactile datasets in the literature is a challenge for studying multimodal fusion strategies analogous to human texture perception. This paper presents a visual dataset that augments an existing tactile dataset. We propose a novel deep fusion architecture that fuses visual and tactile data using four types of fusion strategies: summation, concatenation, max-pooling, and attention. Our model shows significant performance improvements (97.22\%) in surface roughness classification accuracy over tactile only (SVM - 92.60\%) and visual only (FENet-50 - 85.01\%) architectures. Among the several fusion techniques, attention-guided architecture results in better classification accuracy. Our study shows that analogous to human texture perception, the proposed model chooses a weighted combination of the two modalities (visual and tactile), thus resulting in higher surface roughness classification accuracy; and it chooses to maximize the weightage of the tactile modality where the visual modality fails and vice-versa.

\end{abstract}

\section{Introduction}





Texture is a multidimensional psychophysical quantity comprising of 1) roughness, 2) warmness, 3) friction, and 4) hardness \cite{okamoto2012psychophysical}. Human texture perception uses vision and tactile sensing complementary to each other \cite{lederman1986perception}. The visual sensing mechanism extracts global key features, while tactile sensing complements the vision system by providing local surface texture features. However, when visual contact with a surface is lost owing to occlusion or reduced field of view, touch sense allows humans to `see'. Thus, through this multi-modal sensory interaction, we `touch to see' and `see to feel'. 

Research indicates that visual and tactile modalities share information for texture perception \cite{ernst2002humans}. While vision and touch perform better for macro details, touch excels in perceiving micro details. Lay, spacing, and depth can all be used to analyze a surface \cite{leung2001representing, ojala2001texture, iso_21920_1_2021}. Visual sensing is better at understanding lay and spacing, whereas tactile sensing is capable of recognizing all three modes. Visual perception of texture is affected by several factors such as: scaling, rotation, translation, color variance, illumination, and Field of View (FoV); tactile perception is less susceptible to the above factors \cite{luo2018vitac}. Inspired by the human intersensory mechanism of texture perception, similar methods can be developed for the artificial exploration of textures by machines. 

Texture exploration by machines is crucial in an industrial setting because it corresponds to product feel, material finish, and adhesion. In this paper, surface roughness as a subset of texture is investigated. Surface roughness has proven reliable for understanding the effects of corrosion, cracks, irregularities, friction, and wear and tear on a finished product in an industry setting \cite{toloei2013relationship, maiya1975effect}.

Recently, several classification algorithms for texture roughness have been developed. Although the task seems simple to humans, it remains a challenge for machines.  Visual texture classification aims to identify if a given image patch or image belongs to a specific texture category. For classification, conventional methods employ several hand-crafted feature-based approaches, including wavelet packet signatures and rotation- and scale-invariant Gabor features \cite{laine1993texture, arivazhagan2003texture, dong2011wavelet, bianconi2007evaluation, riaz2013texture}. Methods based on Singular Value Decomposition (SVD), Support Vector Machine (SVM), local binary pattern (LBP), and Convolutional Neural Network (CNN) are also explored \cite{selvan2007svd, ojala2000gray, ojala2002multiresolution, guo2010completed, banerji2011novel, fujieda2018wavelet, andrearczyk2016using}. Several studies also analyze texture with tactile sensing technologies such as GelSight (as a tactile sensor) or a whisker sensor \cite{yuan2017gelsight, wolfe2008texture, jadhav2010texture, hipp2006texture, giguere2009surface, dallaire2014autonomous}.  Two studies used maximum covariance to combine both visual and tactile sensing from GelSight \cite{luo2018vitac} and Whisker \cite{ kroemer2011learning} for texture classification. However, since the model lacks the ability to choose between features from different modalities, an inferior feature from a particular modality might result in degradation of the overall model performance. Further, these studies lack the multi-modal mechanism of texture perception of humans primarily due to the lack of standardized visuo-tactile dataset. Finally, fusing of these two modalities for texture classification is relatively less explored.

In this study, we examine the surface roughness of a standard specimen as a test bed for an intersensory information-sharing deep learning model. The tactile sensing can perceive details of micro roughness, while visual sensing can provide a global picture of the specimen. The deep-learning model should be capable of combining features from these two modalities for a more accurate surface roughness classification.

\begin{figure*}[ht]
\centering
\includegraphics[width=\textwidth]{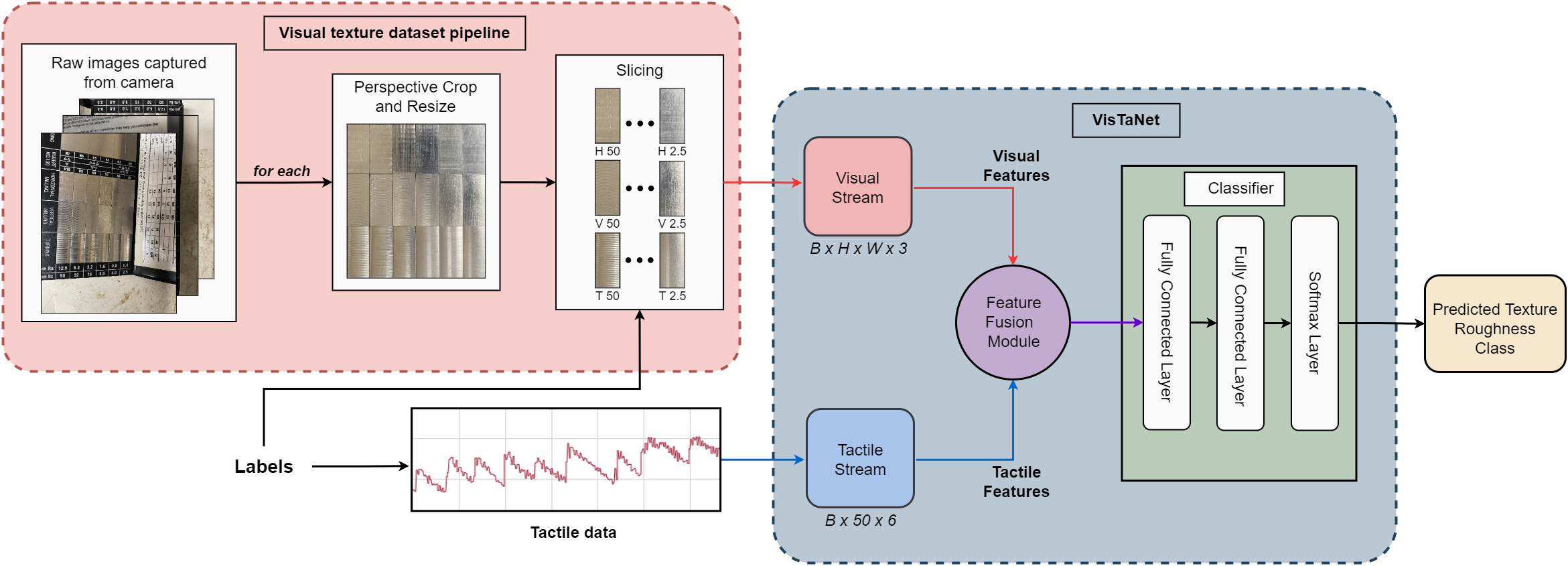}
\caption{\textit{Overview of the visuo-tactile dataset pipeline and the proposed network architecture.} The visual texture dataset pipeline provides the outline for processing the captured images from the specimen. The visual and tactile streams extract features from corresponding visual and tactile data. The feature fusion module combines the two features passing them to the classifier downstream. The classifier module comprises a stack of two fully-connected layers followed by an 18-class softmax layer which predicts the texture roughness.}
\label{fig:mainfigure}
\vspace{-2mm}
\end{figure*}

In this paper, we investigate methods to improve the accuracy of surface roughness classification using deep-learning models. Existing literature suggests that human texture perception can be interpreted as a weighted average of multi-sensory inputs. Achieving similar performance in robots would require a coherent visuotactile dataset. To this end, we propose a visual dataset that augments an existing tactile dataset. Further, we present a novel attention-guided deep fusion architecture that combines the visual and tactile modalities using a weighted average. Our model shows significant performance improvements over \textit{tactile only} \& \textit{visual only} architectures for surface roughness classification. Finally, we perform an ablation study comparing various fusion strategies for combining visual and tactile features showing that the attention-based fusion strategy offers the best performance improvements.

\section{Preliminaries}


We begin by emphasizing the need to combine the visual and tactile modalities for surface roughness classification. We discuss an existing tactile dataset with its associated mathematical notation. Further, we present a high-quality visual dataset that would be useful when paired with the tactile dataset, improving the performance of deep-learning models. Finally, we present a mathematical introduction to the concept of attention, a crucial component of the fusion mechanism used in the proposed model.


\subsection{Need for Multi-Modal Texture Data}
In order to ensure the realization of sensor fusion-based texture classification, it is beneficial to use a coherent visuotactile dataset. Authors in \cite{routray2022towards, campion2005fundamental} identified fundamental limits in capturing texture in the visual and tactile domains. Even though tactile data provides rich surface roughness information, it is limited to a one-dimensional sweep over the specimen using a whisker sensor. On the other hand, visual data captures a two-dimensional representation of the surface. Therefore, combining local information from the tactile data with the global information from the visual data is crucial for surface roughness understanding. We extend the existing tactile dataset with our image dataset, thus filling a critical gap in the literature. Finally, we present a mathematical introduction to the attention mechanism, a crucial building block of \textit{VisTaNet}.

\subsection{Tactile Dataset}
\begin{figure}[h!]
\centering
\includegraphics[width=0.38\textwidth, height=0.3\textheight]{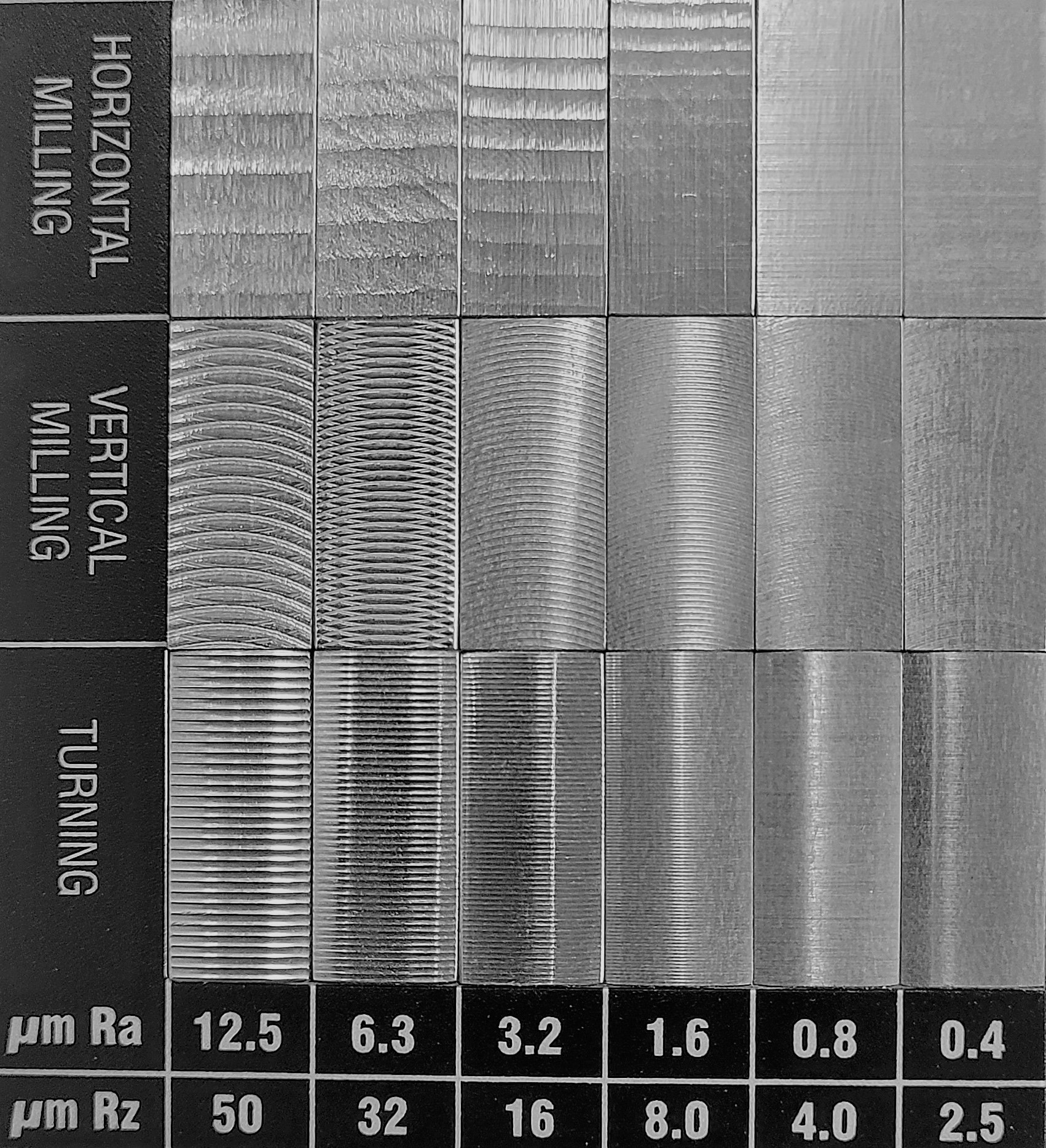}
\caption{\textit{Roughness specimen.} A steel surface machined to get roughness of different depth and spacing. Adapted from \cite{routray2022towards}}
\label{fig:roughnessSpecimen}
\vspace{-2mm}
\end{figure}

For our study, we have used the available tactile dataset collected using a whisker based sensor \cite{routray2022towards} over a standard machined roughness specimen by RUBERT \& Co. LTD., England as shown in Fig. \ref{fig:roughnessSpecimen}. Tactile data of the existing dataset was collected from sweep over specimen at two sweep speeds of $50 \; mm/min$ and $100 \;  mm/min$. We have used the data of sweep speed of $50 \;  mm/min$, since it captures the roughness information at a much higher resolution. Further, we divide the sweep length as follows:

Let $x \in \mathbf{R}^{D}$ represent the D-dimensional feature collected by the sensor for each sample; $D = 6$ for the whisker sensor used in our study (3-axial Accelerometer and 3-axial pressure sensor). The tensor $X = \{x_1, x_2, ..., x_i,... x_T\}$; where $X \in \mathbb{R}^{T \times D}$ and $x_i \in \mathbb{R}^{D}$, represents the time-series data collected over a surface with label $Y$ in a sweep. The sweep data has a large dimensionality, which could pose a challenge for training a machine learning model. Additionally, literature suggests that smaller movements can lead to better understanding of texture roughness as opposed to large movements across the surface \cite{}. Hence, we split the tensor $X$ into temporal windows of small size $W$ such that $X_{i}^{W} = \{x_{i}^W, x_{i+1}^W, ..., x_{i+W-1}^W\}$, where $X_{i}^{W} \in \mathbb{R}^{W \times D}$. The tuple $(X_{i}^W, Y)$ is used for training the deep learning models classifying texture roughness label $Y$ with the input to the model $X_{i}^W$. The window size W is a hyperparameter, and a chosen optimal value is 50.

\subsection{Visual Dataset}
We have acquired RGB images of the standard roughness specimen using a smartphone camera (12 MP, f/1.6, 26mm) positioned vertically on an adjustable camera platform.Fig.\ref{fig:mainfigure} shows the overview of the image texture dataset pipeline. To address anisotropic metal surfaces \cite{westin1992predicting}, we ensured that images were taken from various orientations and lighting conditions. To account for varying orientations during image capture, we perform a perspective crop on each image to align them to the same orientation. For further processing, the images are re-scaled to the same dimensions ($610 \times 278 \times 3$). The individual slices of each image are then mapped (I) to their matching ground labels (Y).

Similar to the convention for ground labels for classes followed by \cite{routray2022towards}, we have the specimen data that includes three classes of surface roughness: abbreviated with `H’ for horizontal milling, `V’ for vertical milling, and `T’ for turning. This notation consistency is also in accordance with the visual data convention. Each class has six roughness sub-classes totaling eighteen classes.

\subsection{Attention Mechanism}
\label{sec:transf-enc}
Transformer architecture has grown in popularity as a solution to many problems, including computer vision, speech, and natural language processing \cite{vaswani2017attention}. The attention mechanism is a fundamental building block of the transformer architecture. The attention mechanism can be viewed as a weighted average of the input where the weights are learned during training. Learnable weighted averaging can help fuse multimodal input streams. We present a brief mathematical discussion on the attention mechanism.

Let $H = [h^{T}_{1}, h^{T}_{2}, ..., h^{T}_{N}]^{T}$ be the input to the attention block where $h_{i} \in \mathbb{R}^{1\times d}$. The inputs to the attention block is projected into $W, Q, V$ by three matrices $W_{K} \in \mathbb{R}^{d \times d_{k}}, W_{Q} \in \mathbb{R}^{d \times d_{k}}, W_{V} \in \mathbb{R}^{d \times d_{v}}$. The self-attention then is calculated as follows.

\begin{equation}
    Q = HW_{Q},\; K = HW_{K},\; V = HW_{V}
\end{equation}
\begin{equation}
    A = \textit{softmax}(\frac{QK^{T}}{\sqrt{d_{k}}}), \; f^{attn}(H) = \textit{softmax}(A)V
\end{equation}

Matrix \textit{A} captures the similarity between queries and keys; the attention matrix is used to generate a rich representation for each position in the input by taking into account its interaction between inputs at different positions. The extension of the attention block to the multi-head attention block used in the proposed architecture is simple and straightforward \cite{vaswani2017attention}. 
\section{Model Architecture}
We design a unique multimodal fusion network for texture roughness classification called VisTaNet. The network fuses the global texture information from visual data with the local tactile information, closely mimicking the human exploration of texture \cite{lederman1982perception}. We discuss each stream individually and alternative fusion strategies. Fig. \ref{fig:mainfigure} presents the proposed architecture for fusing multimodal sensory information toward accurate texture classification.

\subsection{Visual Stream Architecture}
\label{sec:visualStream}
The visual stream derives the global texture information in visual data, as seen in Fig. \ref{fig:mainfigure}, which is part of the two-stream architecture. The main building block in the visual stream is a pre-trained ResNet-$50$ \cite{he2016deep} architecture without the final classification layer. For training purposes, we froze all layers except the last bottleneck layer. We also used an augmentation layer before feeding the images to the ResNet-50 to ensure invariance to conditions such as lighting differences and noise during data capture. We use flip, rotation, contrast, translation, and zoom, which are all random ways to improve an image. We use flip, rotation, contrast, translation, and zoom, in every possible way to improve an image. In a real-world scenario, robots that work close to the surface, such as a wall or obstacle, will not have a wide field of view; hence, a $(224 \times 224)$ random crop of the specimen is taken to ensure a narrower field of view.

Mathematically, the visual stream can be represented as a function ($f_{visual}(.)$) which extracts meaningful features ($\mathbb{F}_{visual}$) from visual data ($I$) for classifying the surface roughness as follows:
\begin{equation}
    \mathbb{F}_{visual} = f_{visual}(I)
\end{equation}

\begin{figure}[ht]
\centering
\includegraphics[width=0.48\textwidth]{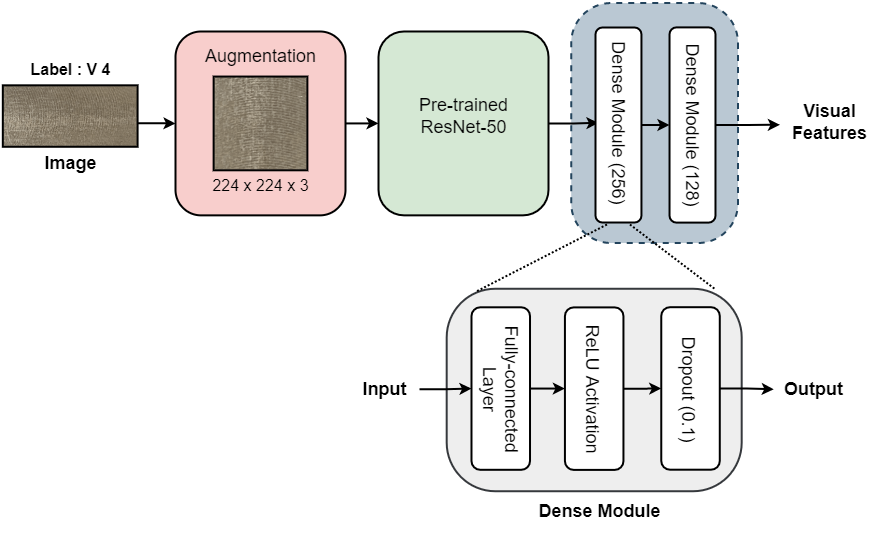}
\caption{\textit{Visual Stream Architecture.} A pre-trained ResNet-50 is the backbone of the Visual stream architecture. The images are passed through an augmentation layer before being fed to the ResNet model. The output of the ResNet is fed to the Dense modules, which extract the visual features and forward them to the Feature Fusion module.}
\label{fig:visualStream}
\vspace{-2mm}
\end{figure}

\subsection{Tactile Stream Architecture}
\label{sec:tactileStream}
\begin{figure}[ht]
\centering
\includegraphics[width=0.48\textwidth]{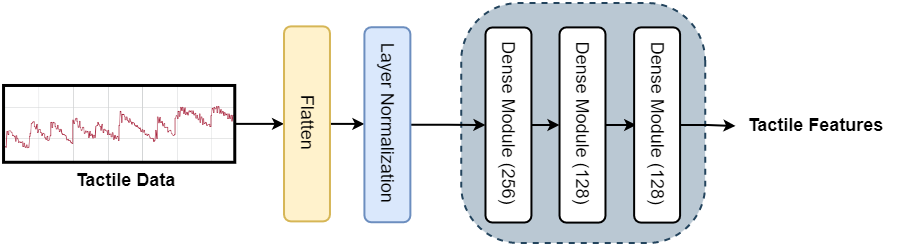}
\caption{\textit{Tactile Stream Architecture.} The tactile data is sent through a Flatten layer followed by the Layer Normalization. It is sent through a series of Dense modules which extract the tactile features and send it to the Feature Fusion module.}
\label{fig:tactileStream}
\vspace{-1mm}
\end{figure}

The tactile stream accounts for extracting texture data locally. The tactile data collected from the whisker sensor resembles humans’ micro-movements during surface exploration. The tactile stream is made up of a layer normalization module followed by a stack of dense modules. Each dense module consists of 1) a fully connected layer, 2) ReLU Activation Layer, and 3) a Dropout Layer, as shown in Fig. \ref{fig:tactileStream}.

Mathematically, the tactile stream can be represented as a function ($f_{tactile}(.)$) which extracts features $\mathbb{F}_{tactile}$ out of the windowed tactile data ($X^W$) useful for classifying the surface roughness as follows:
\begin{equation}
    \mathbb{F}_{tactile} = f_{tactile}(X^W)
\end{equation}

\subsection{Modality Fusion}
\label{sec:fusionMethods}
Humans use a weighted sum voting of visual and tactile information to understand texture \cite{lederman1986perception}. The Attention layer described in Sec. \ref{sec:transf-enc} mimics this phenomena. 
We propose to use the attention layer for fusing the visual and tactile streams to improve texture roughness classification. 

We introduce notations to discuss the fusion mechanism. A fusion function $f^{comb}\; : \; x^a, x^b \longrightarrow y$ fuses the visual stream feature $x^a \in \mathbb{R}^D$ and the tactile stream feature $x^b \in \mathbb{R}^D$. The fused feature $y$ is then passed on to the downstream layers. Fig.\ref{fig:mainfigure} shows the architecture diagram for visuo-tactile surface roughness classification.

\subsubsection{Summing Fusion} $y^{sum} = f^{sum}(x^a, x^b)$ computes the sum of the two features at the same feature dimension $d$:
\begin{equation}
\label{eq:sum_feat}
y^{sum} = x^a_d + x^b_d    
\end{equation}
where $1 \leq d \leq D$ and $x^a, x^b, y \in \mathbb{R}^D$

\subsubsection{Max Fusion} $y^{max} = f^{max}(x^a, x^b)$ computes the maximum of the two features at the same feature dimension $d$:
\begin{equation}
y^{max} = max(x^a_d , x^b_d)
\end{equation}
where all variables are defined as above eq. \ref{eq:sum_feat}.

\subsubsection{Concatenate Fusion} $y^{concat} = f^{concat}(x^a, x^b)$ stacks the the two features on the feature dimension $d$:
\begin{equation}
y^{concat} = concat(x^a_d , x^b_d)
\end{equation}
where $y \in \mathbb{R}^{2D}$, since stacking the two features increases the dimensionality by a factor of 2.

\subsubsection{Attention Guided Fusion}  $(y_a^{attn}, y_b^{attn}) = f^{attn}(x^a, x^b)$ computes a weighted sum of the two features:

\begin{equation}
y_a^{attn} = p x_{a} + q x_b \; , \;
y_b^{attn} = r x_{a} + s x_b
\end{equation}

where $0 \leq p,q,r,s \leq 1 ; \; p + q = 1 \; , \; r + s = 1$; and $y \in \mathbb{R}^{D}$. We only pass $y_a^{attn}$ to the downstream layers. This selection optimizes $p\; \&\; q$ in a way that improves the classification result. The downtream layers of the proposed model consists of a stack of two fully-connected layers with linear activation followed by a $18$ class softmax layer which predicts the class of the surface. 

Mathematically, the proposed architecture which extracts features $(\mathbb{F}_{fused})$ out of the visuotactile input $(I, X^W)$ for classifying surface roughness is defined as follows:

\begin{equation}
    \mathbb{F}_{fused} = f^{comb}(\mathbb{F}_{visual}, \mathbb{F}_{tactile})
\end{equation}
\begin{equation}
    \hat{y} = softmax(\mathbb{W} \times \mathbb{F}_{fused})
\end{equation}

Here, the stack of two fully-connected linear layers is represented by a learnable matrix $\mathbb{W}$; we have omitted the bias term for simplicity. The network is trained using the cross-entropy loss between the ground-truth texture roughness class and the predicted class by the network given by:

\begin{equation}
    \mathcal{L}(y; \theta) = CrossEntropy(y, \hat{y})
\end{equation}

\section{Results and discussion}
\subsection{Setup}
\paragraph{Common Configuration}

All deep-learning models presented in this paper were trained using a common configuration: Nvidia RTX 3090 GPUs were used to train the networks. We trained for $100$ epochs with a batch size of $64$. The Adam optimizer has an initial learning rate of $0.001$ and a decay in learning rate that is exponential, decreasing by a factor of $0.1$ after every $25$ epochs. We used a split of $80:20$ for training and test sets.

\paragraph{Visual texture classifier}
Using image data, we compared the Visual stream (see III-A) texture classifier against the Wavelet-CNN classifier [18] and the FENet-50 classifier \cite{xu2021encoding}. Each architecture has a softmax layer with $18$ classes attached to the last layer.



\subsection{Results}
\label{sec:results}
\subsubsection{Comparative Analysis}
We undertake a comparative analysis between: 1) visual, 2) tactile, and 3) proposed model. The results are presented in Table \ref{tab:AGResults}. The ResNet-$50$ and FENet-$50$ significantly outperform the WaveletCNN \cite{FujiedaTH17} in terms of classification accuracy. Wavelet CNN relies on exploiting repetitive texture patterns via frequency domain analysis, which reflects in their poor performance with the standard specimen of surface roughness. It is important to highlight that FENet-$50$ uses a pre-trained (on ImageNet \cite{russakovsky2015imagenet}) ResNet-$50$ backbone, which could be a reason for almost equivalent performance. Experimentally, we observed a significant drop in accuracy if randomly initialized ResNet-$50$ backbone is used.

\begin{table}[h!]
\centering
\caption{Texture Roughness Classification results}
\begin{tabular}{l|S} 
    \hline
      \textbf{Classifier} & \textbf{Accuracy (\%)  $\uparrow$}\\
      \hline
      \multicolumn{2}{c}{Visual}\\
      \hline
      Wavelet-CNN \cite{FujiedaTH17} & 60.18\\
      ResNet-$50$ \cite{he2016deep} & 83.33\\
      FENet-$50$ \cite{xu2021encoding} & 85.01\\
      \hline
      \multicolumn{2}{c}{Tactile \cite{routray2022towards}}\\
      \hline
      MLP & 80.93\\
      RF & 89.25\\
      SVM & 92.60\\
      \hline
      \multicolumn{2}{c}{Fused modality}\\
      \hline
      VisTaNet (Ours) & \textbf{97.22}\\
      \hline
    \end{tabular}
\label{tab:AGResults}
\end{table}

We observe that the proposed model outperforms visual/tactile data-only models for classifying surface roughness. The improvements in performance are a direct result of combining the visual and tactile modalities, which is comparable to texture perception in humans. 

\subsubsection{Visual Modality vs Tactile Modality - A Study}
\begin{figure}[htb!]
\centering
\includegraphics[width=0.3\textwidth, height=0.25\textwidth]{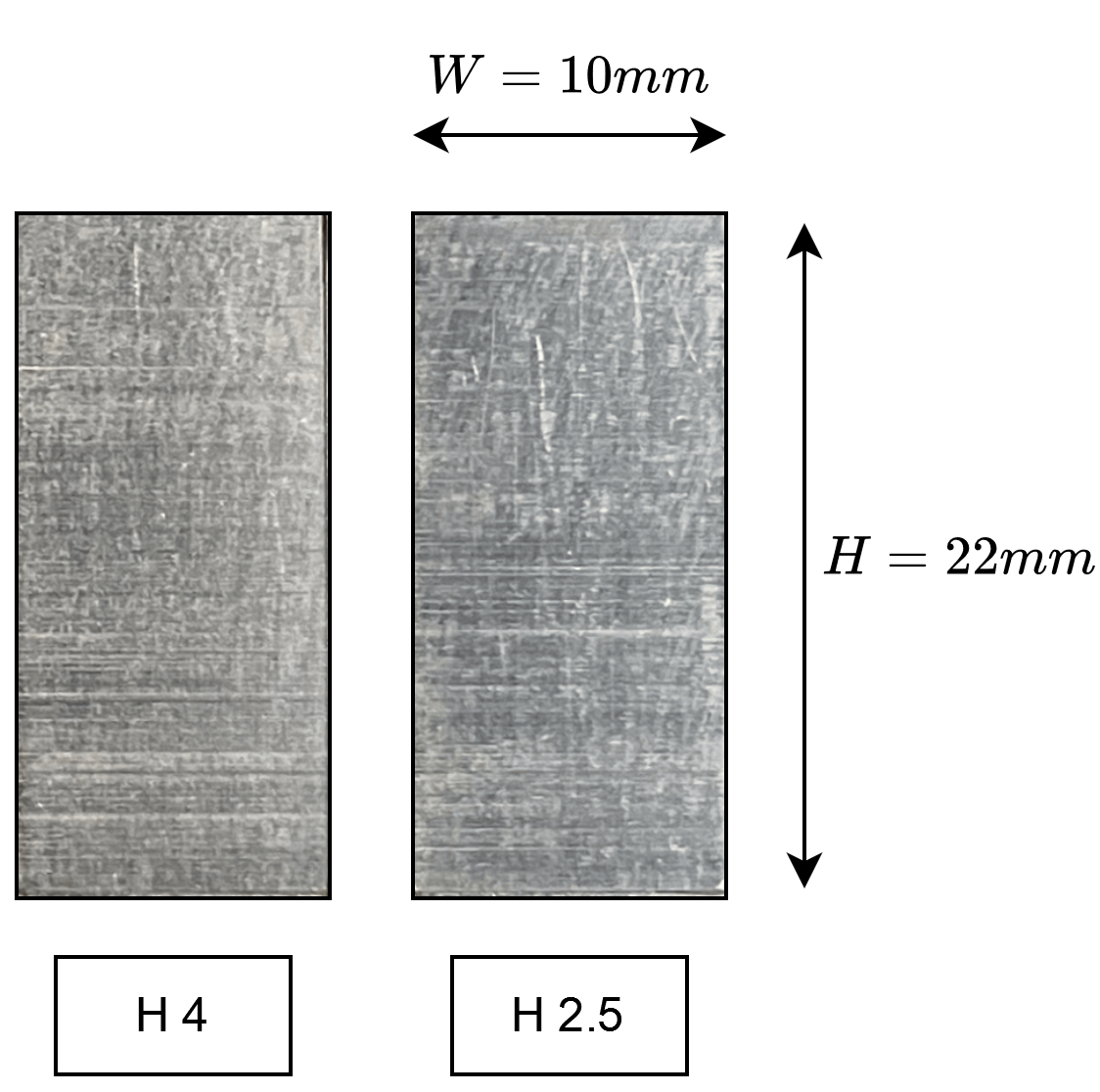}
\caption{\textit{Visually indistinguishable classes.} H 2.5 and H 4 classes of surface look visually similar, which makes the classification process difficult for a visual texture classifier.}
\label{fig:visuallyCloseTexture}
\end{figure}

Fig. \ref{fig:visuallyCloseTexture} illustrates two visually indistinguishable classes of horizontal milling (H-2.5 and H-4). Even for a human expert, it is difficult to identify the difference between the two surface images. The challenge stems from the inherent reduction in resolution while digitizing the surface. A robotic device that uses a vision-based system to differentiate between surfaces dominated by micro-roughness characteristics would encounter a similar challenge. As shown in Fig. \ref{fig:visuallyCloseTexture}, $H_i = 610 $ and $W_i$ = 278 represent the specimen dimensions in terms of pixels in the captured image, while $H$ and $W$ represent the real-world length of the roughness specimen. Therefore, we can find the area each pixel captures in terms of height per pixel ($L_p$) and width per pixel ($W_p$). $L_p = \frac{H}{H_i} = 32.44 \; \mu m/pixel$ and $W_p = \frac{W}{W_i} = 35.97 \;  \mu m/pixel$. $L_p$ and $W_p$ indicates the inadequacy of visual data for the roughness classification challenge. Due to the relatively close spectral information between neighboring classes, the frequency domain processing of texture information degrades the classification results.

\begin{figure}[ht]
\centering
\includegraphics[width=0.42\textwidth]{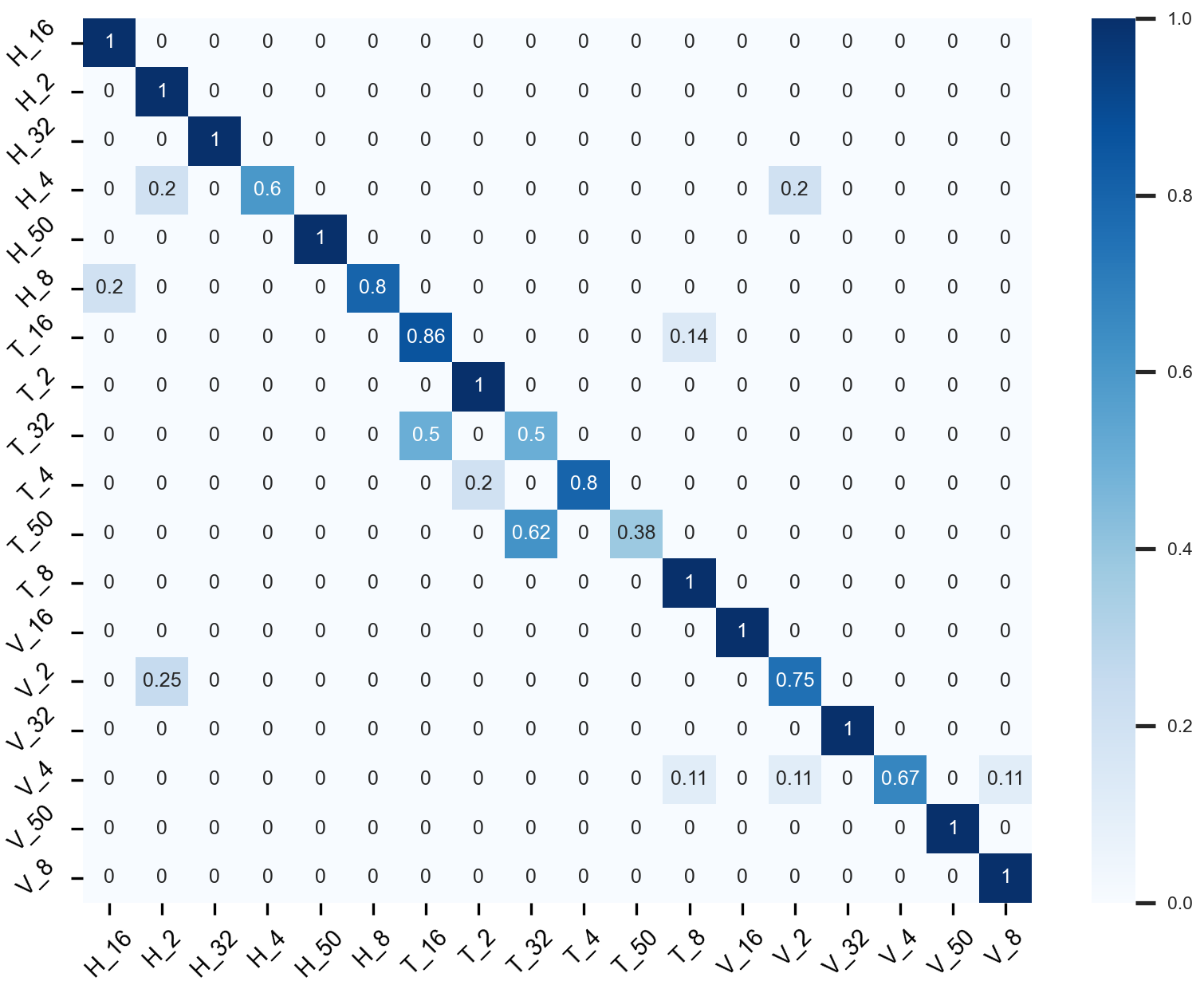}
\caption{Confusion matrix for image-based classifier (ResNet-$50$)}
\label{fig:resnet50VisualStream}
\end{figure}

Fig. \ref{fig:resnet50VisualStream} shows the ResNet-$50$ classifier’s confusion matrix. A Majority of misclassifications fall into either a roughness class with a close-by granularity or a roughness class with the same granularity but a different milling type. Also, most incorrect classifications emerge from finer granularity ( $ \leq 32 \micro m$) since the image data doesn’t have enough resolution to capture these details. Upgrading to a camera of higher resolution would not be a practical choice due to the constraints placed on the robot operating in an industrial setting.

Table \ref{tab:AGResults} and earlier studies show that the simple SVM classifier significantly outperforms the image-only trained models. This is due to a much higher resolution of the surface captured by the whisker sweeps over the surface. Results from \cite{routray2022towards} show that intra and inter-class misclassification rates are lower. Another finding is that the misclassified labels had a wider dispersion. This could be owing to the intrinsic noise the whisker recorded during the stick-slip surface exploration mechanism \cite{wolfe2008texture}. The visual or tactile data are insufficient to provide the best classification results.

\begin{figure}[h!] 
    \centering
    \includegraphics[width=0.42\textwidth]{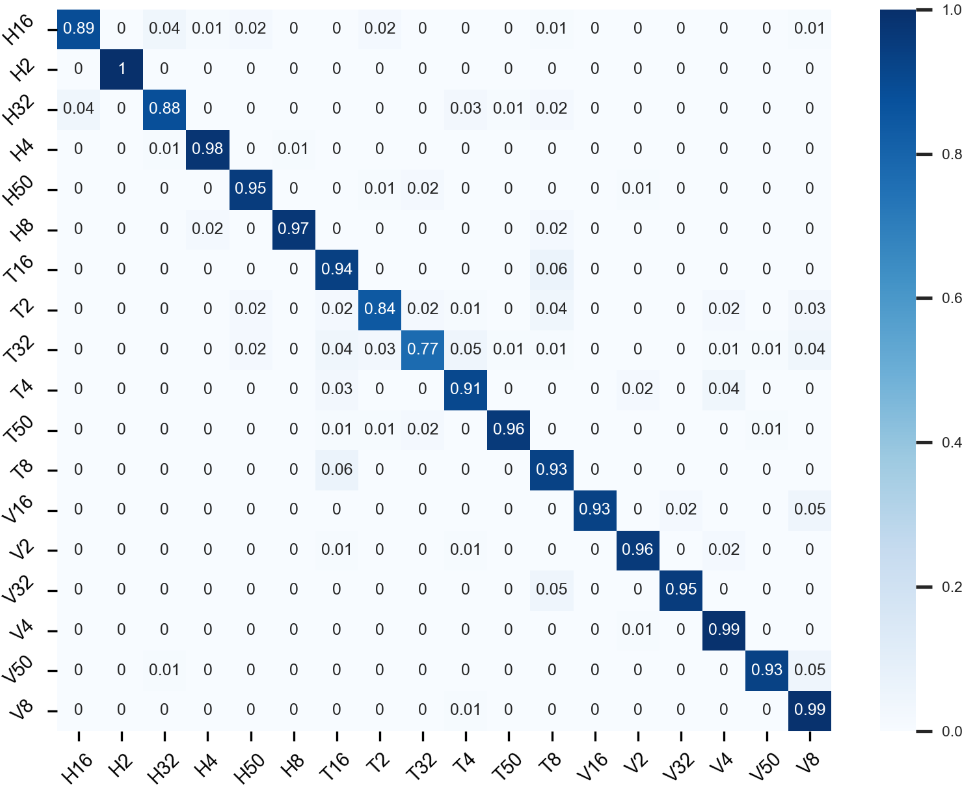}
    \caption{Confusion matrix for SVM calssifiers' performance on texture data.}
    \label{fig:tactileConfusionMatrix}
    \vspace{-1mm}
\end{figure}

\begin{figure}[h!]
\centering
\includegraphics[width=0.42\textwidth, height=0.35\textwidth]{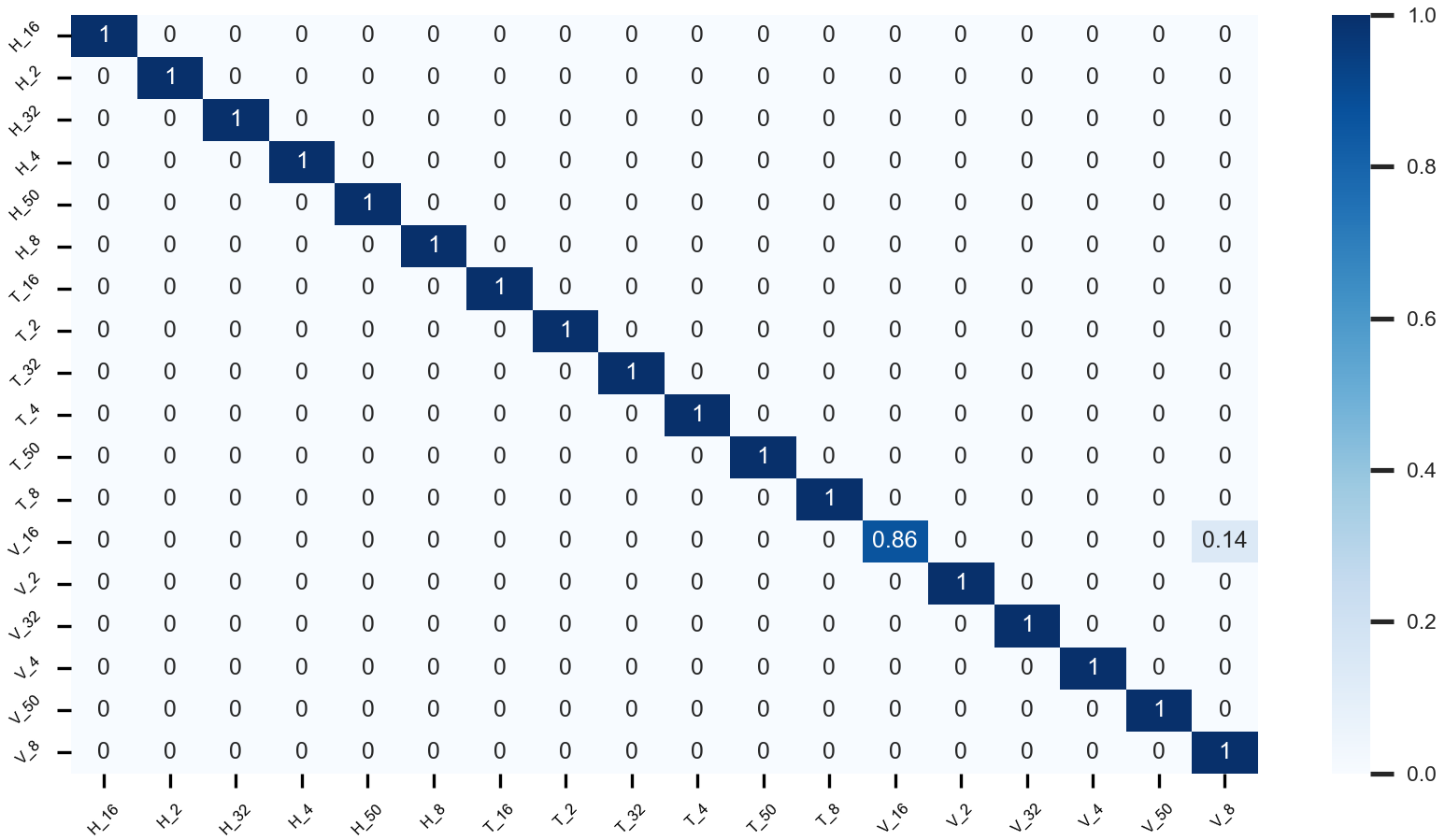}
\caption{Confusion matrix for surface roughness classification by VisTaNet on the dataset.}
\label{fig:AG_confusionMatrix}
\vspace{-1mm}
\end{figure}

\subsection{To Look or to Feel?}

\begin{figure}[h!]
\centering
\includegraphics[width=0.49\textwidth]{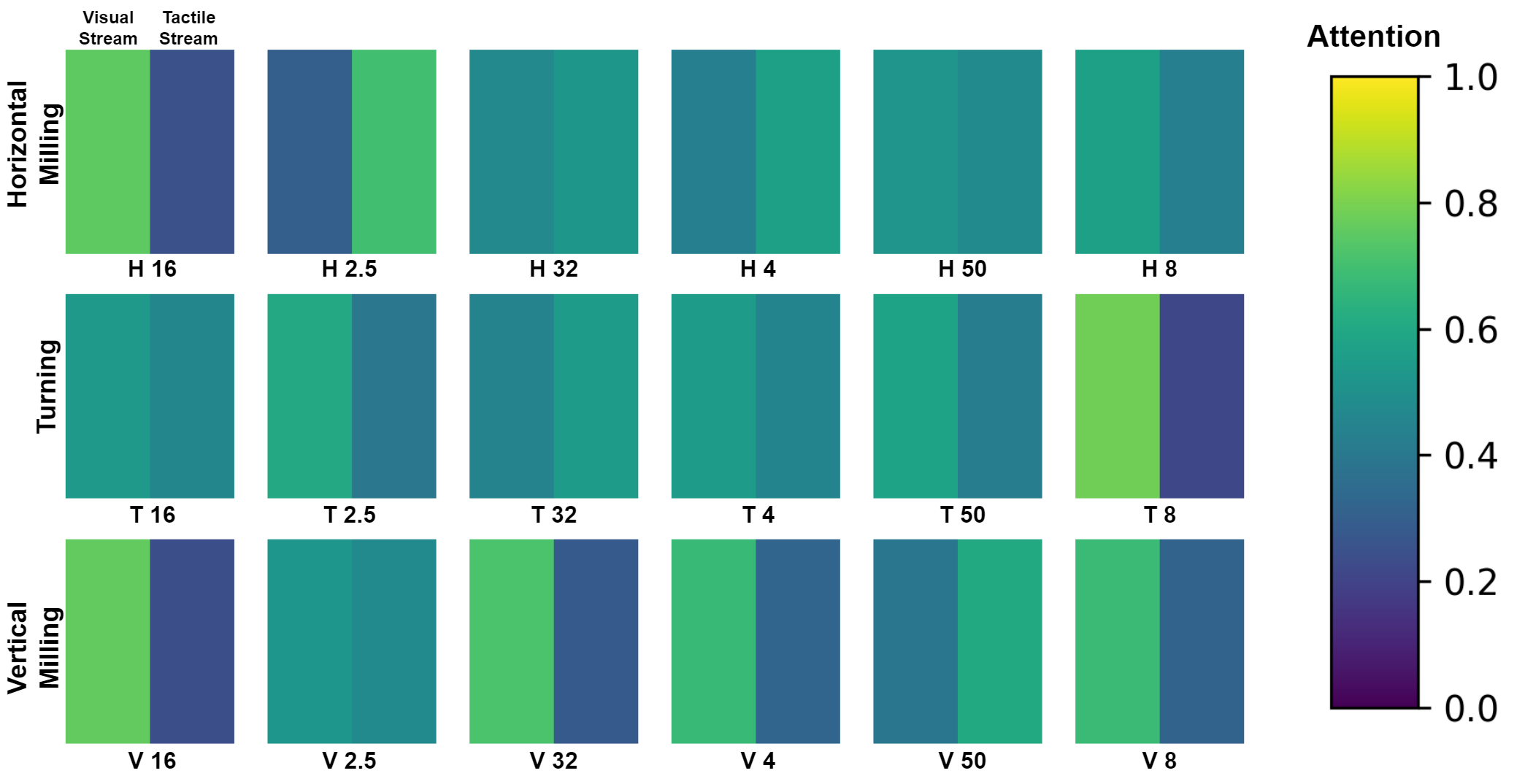}
\caption{\textit{To Look or To Feel.} The plot is divided into 18 subsections with 3 rows and 6 columns. Each row represents a milling type, whereas each column represents a particular granularity. The attention plot for each class is represented by a $(1 \times 2)$ image vector, which contains the weights determined for visual and tactile modality for that particular class. Gradient color-bar on the right can be used to infer weightage distribution between the two modalities.}
\label{fig:attentionPlot}
\end{figure}

We discussed the importance of multi-modal fusion for texture perception in humans. We showed that an analogous version can be achieved in deep-learning models via attention guided fusion mechanism. The attention mechanism calculates a weighted average of the two modalities (visual and tactile). Let $a$ and $1 - a$ ($0 \leq a \leq 1$) represent the weights for visual and tactile stream respectively. The trained model will optimize $a$ \& $1 - a$ in a way that maximizes the classification accuracy. We present the average attention plot for all the 18 classes in Fig \ref{fig:attentionPlot}. The plot is divided into 18 subsections with 3 rows and 6 columns. Each row represents a milling type, whereas each column represents a particular granularity. The attention plot for each class is represented by a $(1 \times 2)$ image vector, which contains the weights determined for visual and tactile modality for that particular class. We have added a gradient color-bar on the right for better understanding of weightage distribution between the two modalities. The use of attention mechanism as a fusion strategy allows the proposed network to mask features coming from a specific modality which might degrade the classification performance.

\subsection{Ablation Study}
We compare the fusion methods discussed in Section \ref{sec:fusionMethods}, evaluating their performance in terms of classification accuracy. The results from Tab. \ref{tab:fusion_strategies} validate using the attention-based fusion of the visual and tactile modalities. The attention plots presented in Fig. \ref{fig:attentionPlot} also make model performance more explainable, since the dominant modality used by the network for classifying the correct texture roughness class can also be inferred. Summation is a non-reversible operation that might combine good and bad features from the two modalities, which could degrade the model performance. In contrast, attention-based fusion enables the network to choose the proportion of features from each modality, thus resolving the issue.

\begin{table}[h!]
    \centering
    \caption{Ablation Study for various fusion strategies used in VisTaNet }
    \begin{tabular}{l|S} 
        \hline
        \textbf{Fusion Strategy} & \textbf{Accuracy (\%)}\\
        \hline
        Summation & 84.72 \\
        Max Fusion & 95.37 \\
        Concatenation & 95.27 \\
        Attention Guided & \textbf{97.22} \\
        \hline
    \end{tabular}
    \label{tab:fusion_strategies}
\vspace{-0.6cm}
\end{table} 

\subsection{Limitation}
The temporal nature of the tactile data is unexploited which could be a major contributing factor towards accurate surface roughness classification. Further, the proposed augmentation of visual dataset with an existing tactile dataset loosely couples the two distinct modalities, this could result in a reduced classification accuracy. 


\section{Conclusion}
The main inspiration for this work was derived from intersensory based texture perception in humans which led us to study the multimodal sensor fusion for surface roughness texture classification. To address the lack of a coupled visuotactile dataset, we introduced a visual dataset which augments an existing tactile dataset. We proposed a deep fusion network which uses the attention mechanism to combine features from visual and tactile modalities. We achieved a classification accuracy of 97.22\% with the proposed model against \textit{tactile only} (SVM - 92.60\%) and \textit{visual only} (FENet-$50$ - 85.01\%) architectures. We observe that most incorrect classifications emerge from finer granularity ($\le$ 32µm) since the image data doesn’t have enough resolution to capture these details. We show that the proposed network shifts attention to the dominant modality, thus closely mimicking human perception. The use of attention mechanism as a fusion strategy allows the proposed network to mask features coming from a specific modality which might degrade the classification performance. 



In future, simultaneous recording of visuo-tactile data may reduce disparity between the data acquired form the same surface. A real-time coherency between 1D tactile data and 2D visual data is required to reduce disparity.




\bibliographystyle{./bibliography/IEEEtran}
\bibliography{./bibliography/refs}

\end{document}